\def\year{2019}\relax
\documentclass[letterpaper]{article} 
\usepackage{aaai19}  
\usepackage{times}  
\usepackage{helvet}  
\usepackage{courier}  
\usepackage{url}  
\usepackage{graphicx}  
\usepackage{times}
\usepackage{latexsym}

\usepackage{url}

\usepackage{subfigure}
\usepackage{times,amsmath,epsfig}
\usepackage{graphicx}
\usepackage{url}
\usepackage{enumitem}
\usepackage{amsfonts}

\usepackage{amsmath}
\usepackage[noend]{algpseudocode}
\usepackage[linesnumbered,ruled]{algorithm2e}
\usepackage{mathtools}

\newcommand{\SAVE}[1]{}

\usepackage{mdwlist}  
\usepackage{multirow}  
\usepackage{amsmath} 
\usepackage{relsize} 

\usepackage{float}
\usepackage{flushend}

\usepackage{graphicx}
\makeatletter
\newcommand*\bigcdot{\mathpalette\bigcdot@{.5}}
\newcommand*\bigcdot@[2]{\mathbin{\vcenter{\hbox{\scalebox{#2}{$\m@th#1\bullet$}}}}}
\makeatother

\usepackage{tikz}

\title{Improving Distantly Supervised  Relation Extraction with Neural Noise Converter and Conditional Optimal Selector
}

\author{Shanchan Wu \\
	Alibaba Group (U.S.) Inc. \\
	{\tt shanchan.wu@alibaba-inc.com} \\\And
	Kai Fan \\
	Alibaba Group (U.S.) Inc. \\
	{\tt k.fan@alibaba-inc.com} \\\And
	Qiong Zhang \\
	Alibaba Group (U.S.) Inc. \\
	{\tt qz.zhang@alibaba-inc.com} \\
}

\date{}

\begin{document}
\maketitle


\begin{abstract}

Distant supervised relation extraction has been successfully
applied to 	large corpus with thousands of relations.	However,
the inevitable wrong labeling problem by distant supervision 
will hurt the performance of relation extraction. 
In this paper,
we propose a method with neural noise converter to alleviate
the impact of noisy data, and a conditional optimal selector to make proper prediction.
Our noise converter learns the structured transition matrix on logit level
and captures the property of distant supervised relation extraction dataset.
The conditional optimal selector on the other hand
helps to make proper prediction decision of an entity pair even if  the 
group of sentences is overwhelmed by no-relation sentences.
 We conduct experiments on a widely used dataset
and the results show significant improvement over competitive baseline methods.

\end{abstract}

\section{Introduction}

The task of relation extraction (RE) is to predict semantic relations between pairs of entities
in text. To alleviate the work load of obtaining the training data for relation extraction, distant supervision is applied to collect training data for learning a model. Relation extraction (RE) under distant supervision  heuristically aligns entities in texts to some given knowledge bases (KBs).
For a triplet $(e_1, r, e_2)$,  where $e_1$ and $e_2$ are a pair of entities with relation type $r$ in KBs, distant supervision
assumes that all sentences
that contain both entities $e_1$ and $e_2$ will express the same relation type $r$ which
will be used in training.

Although distant supervision is an effective way to automatically label training
data, it always suffers from the problem of wrong labeling.
For example, $(Mark Zuckerberg, founder, Facebook)$ is a relational triplet in KBs.
Distant supervision
will regard all sentences that contain
$Mark Zuckerberg$ and $Facebook$  as the instances containing relation $founder$. 
For example, the sentence
\textit{``Mark Zuckerberg uses Facebook to visit Puerto Rico in VR.''}
does not express the relation $founder$
but will still be regarded as a positive instance containing that relation type.
This will introduce noise into the training data.

Tackling the noisy data problem for distant supervision is a non-trivial problem,
as there is not any explicit supervision for the noisy data.  Some previous work
tries to clean the noisy training data.
\citeauthor{Shingo_ACL_2012} propose a solution to identify those
potential noisy sentences and remove them from training data
by  syntactic patterns during the preprocessing stage.
Although the effort of removing noise during preprocessing is practically effective,
it has to rely on manual rules and hence is unable to scale.
Some other work directly makes effort to reduce the impact of 
the noisy data during the training process. 
\citeauthor{Riedel_ECML_2010} make the \textit{at-least-one} assumption 
that
%
if two entities participate in a relation, at least one sentence that
mentions these two entities might express that relation.
They then propose a graphical model to build relation classifier.
There are some following works based on this assumption
using  probabilistic graphic models 
\cite{Hoffmann_ACL_2011,Surdeanu_emnlp_2012},
and neural network methods  \cite{Zeng_emnlp_2015}.
The methods based on \textit{at-least-one} assumption can reduce the impact of some noisy 
sentences. However, it can also lose a large amount
of useful information containing in neglected sentences.
\citeauthor{Lin_ACL_2016}
propose to use sentence-level selective attention mechanism
to reduce the noise through a sentence bag. The selective attention 
mechanism represents a group of sentences including the same entity
pair into a vector representation, and then use it for classification.
This method has very significant improvement over other previous methods.
The selective attention mechanism uses the information of all sentences in the group 
containing the entity pair. Moreover, it still considers that the label of the 
group of sentences in the training data is always correct, and it is further used to
derive the loss
function. However, under distant supervision, it is very likely that
none of the sentences in the same sentence group with the same entity pair express 
the semantic relation decided by KBs.
Neither \textit{at-least-one} solution nor selective attention solution is able to handle
this situation properly.


In this paper, we propose a model with neural noise converter and conditional optimal selector
for distant supervised relation extraction to tackle the noisy data problem. 
We regard the relation label for a pair of entities decided by KBs as a noisy label
for the corresponding group of sentences  obtained by distant
supervision. We build a neural noise converter which tries to build the connection
between true labels and noisy labels. Our final prediction output is then based on the
prediction of true labels rather than noisy labels using a conditional optimal selector.
Some previous works in computer vision such as  \cite{Sukhbaatar_ICLR_2015} have used a ``noise'' layer on top of the softmax layer to adapt  the softmax 
output to match the noise distribution. Our noise converter is different from theirs
as our noise converter is on top of logits rather than softmax, and we 
take advantage of the properties in
the distantly supervised relation dataset and impose a reasonable constraint to the transition matrix. 
One advantage is that we do not need to project the
transition matrix to the space representing  a valid probability distribution for each update. 
Another advantage is that 
the linear transition on hidden vectors
can achieve non-linear transition results
on probability space.
Our noise converter can also theoretically 
converge to its optimal value.
The conditional optimal selector on the other hand
helps to make proper prediction decision of an entity pair even if  the 
corresponding group of sentences are overwhelmed by no-relation sentences.

Our main contributions in this paper are: 
(1)We propose a novel model named neural noise converter to better deal with noisy 
relational data. (2) We design a conditional optimal selector to help make proper
prediction among sentence bags.
(3)Our experimental results demonstrate the effectiveness of our approach over
the state-of-the-art method.

\subsection{Related Work}  
\label{related}

In recent years, convolutional neural networks
(CNN) have been successfully applied to 
relation extraction
\cite{Santos_ACL_2015,Nguyen_NAACL_2015}. 
Besides the typical CNN, variants of CNN have
also been developed to relation extraction, such as 
piecewise-CNN (PCNN) 
\cite{Zeng_emnlp_2015},
split CNN 
\cite{Adel_naacl_2016},
CNN with sentencewise
pooling \cite{Jiang_Coling_2016}
and attention CNN 
\cite{Wang_ACL_2016}.
Furthermore, recurrent neural networks
(RNN) are another successful choice for relation extraction, such as recurrent
CNNs 
\cite{Cai_ACL_2016}
and attention RNNs 
\cite{Zhou_ACL_2016}.


For relation extraction, the effect of wrongly labeled 
training data has been attracted attention 
and some recent works have been proposed
to reduce the influence. 
Among the works, some of them attempt to clean the noisy training data, such as
\cite{Shingo_ACL_2012}. 
The work in \cite{Xu_ACL_2013} 
tries to find possible false negative data through
pseudo-relevant feedback to expand training data.
Some other works directly make effort to reduce the impact of 
the noisy data during the training process,
such as
\cite{Riedel_ECML_2010,Hoffmann_ACL_2011,Surdeanu_emnlp_2012,Zeng_emnlp_2015}. 
%
%
%
\citeauthor{Lin_ACL_2016}
propose to use instance-level selective attention mechanism
to reduce the noise through a sentence bag. The approach
has significantly improved the prediction
accuracy for several baselines of deep learning models.
%


Out of the NLP domain, in computer vision, recently there are  some proposed methods
to fit the noisy data.
\citeauthor{Reed_CoRR_2014} propose a bootstrapping mechanism 
by augmenting the  prediction objective with a concept of perceptual
consistency to make the model more robust to noise. 
\citeauthor{Sukhbaatar_ICLR_2015} 
add a linear “noise” layer on top of the softmax layer which adapts the softmax 
output to match the noise distribution.  
The main issue in \cite{Sukhbaatar_ICLR_2015} is that
the parameters of the extra layer in their proposed model
is not identifiable, and they circumvent this problem by using
a tricky optimization process. 
Rather than optimizing a global transition matrix,  
\citeauthor{Misra_CVPR_2016}
generate the transition matrix in an amortized way for each training instance.
%
%
%
In NLP, as far as we know, the only couple of works that are based on 
neural network noise model are
\cite{Fang_CONLL_2016} and \cite{Luo_ACL_2017}.
The key difference is
that we do not follow the previous work by directly
adapting the softmax output. Instead, we creatively make transition on logits output
and we address the identifiable issue by adding extra constraint to the transition matrix. 

\section{Methodology}  
\label{sec:methods}


\begin{figure*}[tb]
	\centering
	\includegraphics[width=1.0\linewidth]{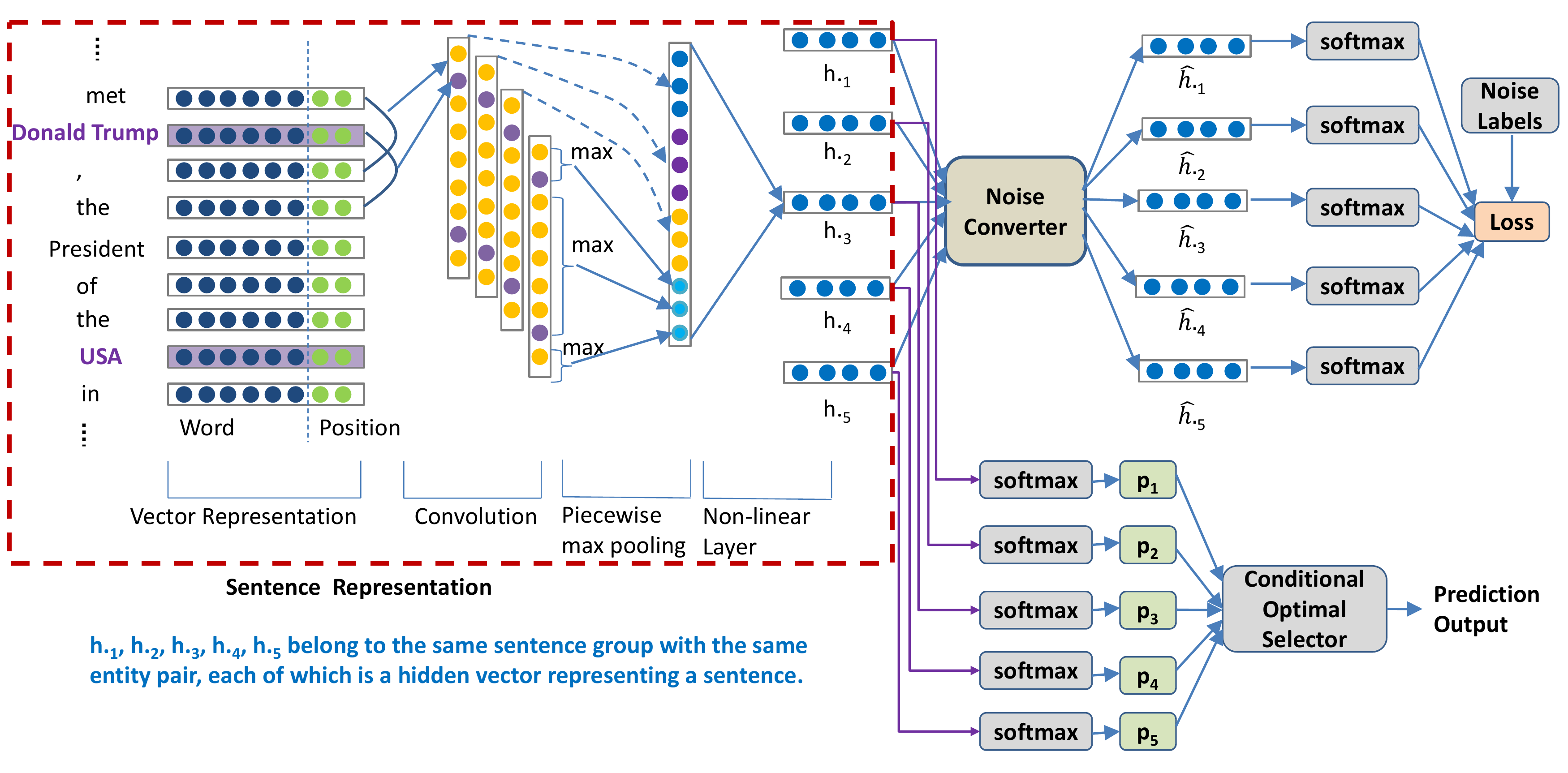}
	\caption{The model architecture. The left side shows the sentence encoder by PCNN. The example
		sentence is encoded to a hidden vector $h_{\bigcdot 3}$.  Four other sentences with the same entity 
		pair are encoded as $h_{\bigcdot 1}, h_{\bigcdot 2}, h_{\bigcdot 4}, h_{\bigcdot 5}$. The noise
		converter converts each hidden vector. Softmax and loss function are applied thereafter.  The conditional optimal selector is used to 
		make prediction based on the hidden vectors which the noise converter is applied to.
	} 
	\label{fig:model_arch}
	\vspace*{-0.5cm}
\end{figure*}


Given a set of sentences $\{s_1, s_2, \cdots· , s_N\}$, with each containing 
a corresponding  pair of entities, our model  predicts the relation type $r$
for each entity pair.  Figure \ref{fig:model_arch} shows the overall
neural network architecture of our model. It primarily consists of three parts:
Sentence Encoder Module, Neural Noise Converter Module, and Conditional Optimal Selector Module.  
Sentence Encoder uses a variant of convolutional neural network (CNN) to 
process a given sentence with a pair of target entities  to a vector representation.
After constructing the vector representation of the sentence, Neural Noise Converter Module converts the hidden state 
with respect  to the true label to the hidden state with respect to  the noise label.
Conditional Optimal Selector  is a mechanism to make prediction 
based on a group of
sentences with the same entity pair. We describe these parts in details below.

\subsection{Sentence Encoder}  

\medskip

Similar to \cite{Zeng_emnlp_2015,Lin_ACL_2016,Ji_AAAI_2017}, we transform a sentence
into its vector representation by piecewise CNN (PCNN,  
a variant of CNN). First a sentence is transformed into a matrix with word 
embeddings and position embeddings.  
Then a convolutional
layer, max-pooling layer and non-linear
layer are used to construct a vector
representation of the sentence. 

{ } { } \textbf{Word Embeddings and Position Embeddings} 
Word embeddings are distributed representations 
of words, which are usually pre-trained from a text
corpus and can capture syntactic and semantic meanings
of the words.  
%
Similar to \cite{Zeng_coling_2014}, we use the word
position information besides the word embeddings. 
The position information has shown its importance 
in relation extraction.  The position feature values of a word
are determined by the distances between this 
word and the two target entity words.
Each relative position value is encoded as a vector which
is randomized at the beginning and updated during the training. 
Then the word embeddings and position embeddings are concatenated, and a sentence is then originally encoded 
as a vector sequence $X = \{x_1, x_2, \dots, x_{|X|}\}$ ,
where $x_i \in \mathbb{R}^d (d=d ^a + d^b \times 2)$, $d^a$ is the dimension of a word embedding, and $d^b$ is the dimension
of a position embedding.  

{ }{ } \textbf{Convolution}
Suppose matrix $A$ and $B$ has the same dimension $t_1 \times t_2$, then the convolution
operation between $A$ and $B$ is defined as $A \otimes B = \sum_{i=1}^{t_1} \sum_{j=1}^{t_2} A_{ij}B_{ij}$ 

Given an input vector sequence $X = \{x_1, x_2, \dots,x_{|X|}\}$ of a sentence,
where $x_i \in \mathbb{R}^d$ represents the vector of the $i$-th word in the sentence. 
Let $X_{i:j} = [x_i : x_{i+1} : \dots : x_j ]$ be a matrix by concatenating the vectors from the
 $i$-th vector  to the $j$-th vector, and let 
 $U = \{U_1, U_2, ..., U_m\}$ be a list of filter matrixes, where the
 $t$-th filter matrix is $U_t \in \mathbb{R} ^ {l \times d}$.
 The convolution operation output between $X$ and $U_t$  will be a new vector 
 $c_t \in \mathbb{R}^{|X|-l+1}$, the $i$-th element of which is
  $c_{ti} = U_t \otimes X_{i:(i+l-1)}$.  Through the convolution operation between $X$ and the list
  of filter matrixes 
  $U = \{U_1, U_2, ..., U_m\}$, we get
  a list of output vectors
  $C = \{c_1, c_2, \dots, c_m\}$.

\begin{figure}
	\centering
	\includegraphics[width=1.0\linewidth]{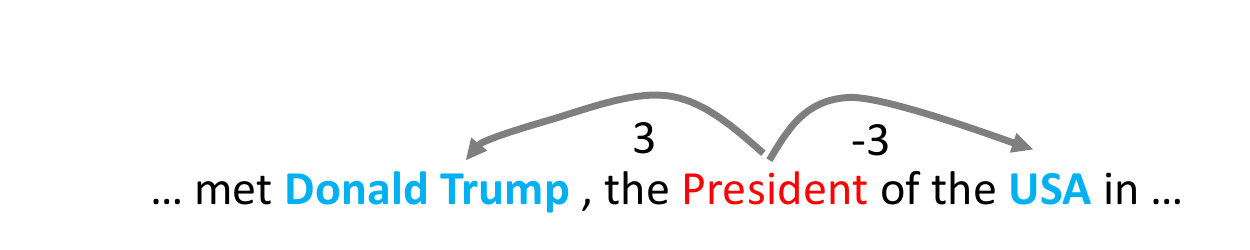}
	\caption{An example of relative distances}
	\label{fig:pcnn}
\end{figure}

 { }{ } \textbf{Max-pooling and non-linear layer}
The max-pooling operation is a widely used method to
extract the most significant features inside a feature map.
To capture the structure information, PCNN divides 
a sequence into three parts based on the positions of
 the two target entities which cut the sequence into
 three pieces. Then the piecewise
max pooling procedure returns the maximum
value in each subsequence instead of a single
maximum value. As shown in Figure \ref{fig:pcnn},
the output vector $c_i$ from convolution operation
is divided into three segments, $\{c_{i1}, c_{i2}, c_{i3}\}$
by \textit{Donald Trump} and \textit{USA}. The piecewise
max pooling procedure is applied to each of the three
segments separately:

$p_{i1} = max(c_{i1}), p_{i2} = max(c_{i2}), p_{i3} = max(c_{i3})$

So for each convolution filter, we can get a vector 
$p_i = \{p_{i1}, p_{i2}, p_{i3}\}$  with 3 elements.
After that, we can concatenate the vectors from
$p_1$ to $p_m$ to obtain the final max pooling output
vector $P \in \mathbb{R}^{3m}$. 

We then apply a non linear layer on the max pooling
output. We also apply dropout on this layer for regularization
during training.

\subsection{Neural Noise Converter}  

\medskip

For the training data obtained from distant supervision, their labels are usually noisy or incorrect. 
To incorporate the noise information, we propose a neural noise converter module that can capture
the relationship between the underlying true labels and the noisy labels. 

We denote $y^*$ and $\tilde{y}$ to be the true label and the observed label (i.e., the noisy label) of the sentence $s$, where $y^*, \tilde{y} \in \{1, \dots, K\}$. 
In addition, we define two probabilistic models $p(y^*|s)$ and $p(\tilde{y}|s)$ to represent the distributions of true and noisy labels, respectively. 
For notation simplicity, we abuse using $p(y^*=k)$ as the alternative of $p(y^*=k|s)$. 

A natural way to convert the true label to the noisy one is to build a linear transformation by assuming the existence of an optimal probabilistic transition matrix $Q^*=(q_{uv}^*)\in R^{K\times K}$ where $ q^*_{uv} =  p(\tilde{y} = u | y^* = v)$ \cite{Sukhbaatar_ICLR_2015}. 
That is to say we can build the following connection 
\begin{equation} \label{eq0}
p(\tilde{y}=u)=\sum_{v=1}^K q_{uv}^* p(y^*=v) .
\end{equation} 
However, we extend it to an implicit transformation by a neural noise converter module, via enforcing a linear transformation on the softmax logits which are used to calculate the relation type distribution. 

To make it concrete, we define $p(y^*) \sim \text{softmax}(h^*)$, where the hidden vector $h^*$ is the softmax logits for true labels. 
Similarly, we have $p(\tilde{y}) \sim \text{softmax}(\tilde{h})$ and the hidden vector $\tilde{h}$ for noisy labels.
In other words, we have
\begin{equation} \label{eq1}
\begin{split}
&p(y^*=k) = \frac{\exp(h^*_{k})}{\sum_{j=1}^{K}\exp(h^*_{j})} \\
&p(\tilde{y}=k) = \frac{\exp(\tilde{h}_{k})}{\sum_{j=1}^{K}\exp(\tilde{h}_{j})}
\end{split}
\end{equation}
Unlike the explicit transformation in probability space \cite{Sukhbaatar_ICLR_2015}, we assume an optimal logits transition matrix $W^* = (w^*_{ij}) \in R^{K\times K}$, such that 
\begin{equation} \label{eq2}
\tilde{h} = W^*h^* .
\end{equation}
A special case is that if $W^*$ is an identity matrix, then it means there is no noise in the observed data. 
Equivalently, we can rewrite Equation (\ref{eq1}) and (\ref{eq2}) to obtain the following approximate linear transformation (due to the presence of $\tilde{Z}$  and $Z^*$ in Equation (\ref{y_to_y}), it is not exact linear) in the log-space of probability,
\begin{equation} \label{y_to_y}
\log p(\tilde{y}=k) = \sum_{j=1}^K w_{kj}^* (\log p(y^*=j) +\log Z^*) -\log \tilde{Z}
\end{equation}
where $Z^*$ and $\tilde{Z}$ are two normalizing terms in Equation (\ref{eq1}). 
In this case, we can not analytically provide the total probability rule as Equation (\ref{eq0}), but we actually define it in the implicit way. 
Notice that the biggest issue in \cite{Sukhbaatar_ICLR_2015} is the identifiable problem of $Q^*$, which also appears in the above definition of $W^*$, since there are $K^2$ free parameters involving only $K$ equations. 
To make our model robust and well-defined during optimization, we suggest imposing some constraint on $W^*$ according to specific tasks. 

\subsection{Structured Transition Matrix}

In this section, we take advantage of the properties in the distantly supervised relation dataset to impose the restriction on the transition matrix $W^*$. 
When two entities are labeled by some relation type, there should be some sentences with that two entities showing that relation type somewhere in the world, but not necessarily in the sentences extracted by distant supervision. 
Meanwhile, that pair of entities in the extracted sentences will rarely show relation types other than `no-relation' or the types that are assigned to this pair from the knowledge base used for distant supervision. 
So the transition matrix should primarily transfer `no-relation' (which is labeled as `1', or negative relation) to positive relations (labeled other than `1'), but unlikely between positive relations or from positive relations to `no-relation'. 
Hence, in the situation of this distant supervised relation extraction task, the transition matrix is assumed in the following form in Equation (\ref{w_restrict}), where the values in diagonal are all 1 except the first one, the rest values except those in the first column and the diagonal are all 0s.
\begin{equation} \label{w_restrict}
\begin{split}
W^* =
\begin{bmatrix}
w_{11}^* & 0  & 0 & \dots  &0 \\
w_{21}^* & 1  & 0 & \dots  & 0 \\
w_{31}^* & 0  & 1 & \dots  & 0 \\
\vdots & \vdots  & \ddots & \vdots \\
w_{K1}^* & 0  & 0 & \dots  & 1
\end{bmatrix}
\end{split}
\end{equation}
Besides the identifiable issue, another purpose to introduce the structured constraint is to guarantee the softmax operation invertible with respect to $W^*$ when $h^*$ is given.  

\subsection{Transition Matrix Estimation}

Let $\hat{p}(y^*|\theta)$ be the prediction probability of true labels by the classification model parameterized by $\theta$. 
Therefore, we can build an extra affine layer with the parameter $W$ (same structure as $W^*$) on top of the softmax logits layer of the true label prediction, converting the distribution of the true labels to the distribution of the noisy labels, i.e., $\hat{p}(\tilde{y}|\theta, W)$. 

Due to the mechanism of distant supervision, the collected dataset are naturally divided into several groups with the same observed entity pairs. 
Thus we can define a customized loss function for the combined noisy model by maximizing the cross-entropy between the observed noisy labels and the model prediction of $\hat{p}(\tilde{y}|\theta, W)$ for the sentences in each group.
The equivalent loss function to minimize is
\begin{equation} \label{loss_def}
\mathcal{L}(\theta, W) = - \frac{1}{G}\sum_{g=1}^{G} \frac{1}{|\mathcal{S}_g|} \sum_{n_g=1}^{|\mathcal{S}_g|} \log \hat{p}(\tilde{y} = r_g | s_{n_g}, \theta, W)
\end{equation}
where $G$ is the total number of entity pairs in the training data, $r_g$ is the observed noisy label in the sentence group $\mathcal{S}_g$ which contains entity pair $E_g$, and $s_{n_g}$ is the sentence included in $\mathcal{S}_g$.

Our purpose is to learn a model such that $W$ is close the underlying optimal $W^*$. Let $\hat{h}$  be the model output of the hidden vector before softmax operation for the noisy labels, and $h$ be the model output of hidden vector before softmax operation for the true labels. 
Similar to Equation (\ref{eq2}) we have $\hat{h} =W h$, thus resulting the following categorical distribution.
\begin{equation} \label{noise_p_express}
\begin{split}
\hat{p}(\tilde{y} = k | s, \theta, W ) &= \frac{\exp(\sum_{j=1}^{K} w_{kj} h_{j})}{\sum_{i=1}^{K} \exp(\sum_{j=1}^{K} w_{ij} h_{j})} \\
	&= \frac{\exp(w_{k1}h_1 + h_k\mathbb{I}_{k\neq1})}{\sum_{i=1}^K \exp(w_{i1}h_1 + h_i\mathbb{I}_{i\neq1})}
\end{split}
\end{equation}
By the above Equation (\ref{noise_p_express}), we can further reformulate the loss function and derive a non-negative lower bound,
\begin{equation} \label{loss_optimize}
\begin{split}
\mathcal{L}(\theta, W) 
=& \frac{1}{G}\sum_{g=1}^{G}  \frac{1}{|\mathcal{S}_a|} \sum_{n_g=1}^{|\mathcal{S}_g|} 
\Bigg( \log \sum_{i=1}^{K} \exp\Big(\sum_{j=1}^{K} w_{ij} h_{j,n_g}\Big) \\
&-  \sum_{j=1}^{K} w_{r_g j} h_{j,n_g}    \Bigg) \\
\ge & \frac{1}{G}\sum_{g=1}^{G}  \frac{1}{|\mathcal{S}_g|} \sum_{n_g=1}^{|\mathcal{S}_g|} 
\Big( \underset{1 \leq i \leq K}{\max}\left( w_{i1} h_{1,n_g} + h_{i,n_g}\mathbb{I}_{i\neq1}\right) \\
& - w_{r_g 1} h_{1,n_g} - h_{r_g,n_g}\mathbb{I}_{r_g\neq1} \Big) \ge 0 
\end{split}
\end{equation}
%
%
where $h_{j,n_g}$ is the $j$th element of the hidden vector for the sentence $s_{n_g}$. 
Based on Equation (\ref{loss_optimize}), when $G \rightarrow \infty $ and $|\mathcal{S}_g| \rightarrow \infty $, an ideal optimization of $\mathcal{L}(\theta, W)$ (i.e., loss approaches 0) will push $w_{r_g 1} h_{1,n_g} + h_{r_g, n_g}\mathbb{I}_{r_g\neq1}$ towards the maximum $w_{i1} h_{1,n_g} + h_{i,n_g}\mathbb{I}_{i\neq1}$ among all entity pairs, for $ \forall  s_{n_g} \in \mathcal{S}_g$. 
It implies that the distribution of the noisy label for each sentence can be calculated after an ideal optimization. 
However, we only care about the distribution of underlying real label during inference, i.e., $\hat{p}(y^*|\theta)$. 
 
In \cite{Sukhbaatar_ICLR_2015}, the total probability rule of the predicted model is explicitly defined in $\hat{p}(\tilde{y})=Q\hat{p}(y^*)$, which implicitly defined via neural noise converter. 
By a deliberate optimization strategy, $Q$ can tend to converge to the optimal unknown $Q^*$ \cite{Sukhbaatar_ICLR_2015}. 
Using this argument, we want to show that our transition matrix $W$ on hidden vectors can also converge to the true transformation matrix $W^*$ by bridging $W$ and $Q$. 
The total probability rule is always hold, thus we have
\begin{equation} \label{p_total}
\hat{p}(\tilde{y}) = \text{softmax}(Wh) = Q \hat{p}(y^*) = Q \text{softmax}(h)
\end{equation}
By the constraint of the structured transition matrix, the softmax operation is invertible with respect to $W$, i.e., given $h$, the solution of $W$ for Equation (\ref{p_total}) is unique. 
Notice the right side $Q \hat{p}(y^*)$ is a $K$-dimensional vector, while the constraint $W$ has only $K$ free parameters. 
As long as $Q$ is able to converge to the optimal $Q^*$, our structured transition matrix can converge to $W^*$. 
 
\subsection{Conditional Optimal Selector}
 After the model is trained, our prediction is then based on the hidden vectors which are the input to
 the neural noise converter. For each pair of entities, 
 a typical solution is to represent the  
 group of sentences with that pair of entities as a single vector by some 
 attention mechanism  and then apply softmax to get the prediction output. 
 However, it is difficult for a single vector representation to make correct prediction
 for a group overwhelmed by `no-relation' sentences if only very few sentences with positive relations are included. 
In this situation, the solution for such a label-imbalance group  will tend to make `no-relation' prediction.
However, the correct label for the target entity pair should be the positive relation type 
expressed in those few sentences.
 
 Instead of directly computing the label distribution of the group with one single vector representation, we first consider to derive all label distributions over sentences, and then select the most representative one for the group label. 
 The selection criteria is basically based on the portion of predicted `no-relation' type of all sentences within each group.
 Thus, we call it ``Conditional Optimal Selector". 
 For a pair of entity, if all sentences in the group are predicted to be negative (i.e. `no-relation' type), we  make `no-relation' prediction for the entire group. 
 Otherwise, if any sentence is predicted to be some positive relation, we will predict the group label based on the sentences with positive relation predictions, regardless of the `no-relation' sentences. 
To be precise, the probability distribution of the group label can be formally expressed as follows.
%
%
%
 \begin{equation} \label{eq:weight}
 \begin{split}
&\text{if }  \forall  s_i , \text{ } \operatorname*{arg\,max}_k \hat{p} (y^*_i = k| s_i \in \mathcal{S}_g) = 1:    \\
&\text{~~~~~~}  j=\operatorname*{arg\,max}_i \hat{p} (y^*_i = 1| s_i \in \mathcal{S}_g)
 \\
&\text{else}: \\
&\text{~~~~~~} j, \kappa=\operatorname*{arg\,max}_{(i, k), k\neq 1} \hat{p} (y^*_i = k| s_i \in \mathcal{S}_g) \\
&\hat{P}(E_g) =   \hat{p}(y_j^*|s_j), 
\end{split}
\end{equation} 
 Furthermore, one side-product is that the multiple positive relation types can also be learned as the final prediction if the task of multiple labels learning is required.
  

\subsection{Model Training}    

We first pretrain the model by replacing $W$ 
with an identity matrix for several epochs,
and then we set $W$ to trainable parameters.

Since we have fixed value 1 in the diagonal of $W$ (See Equation (\ref{w_restrict}) ),
we initialize the variables in the first column in the way with their summation 
equal to 1, in order to make all elements in the matrix at the same scale.

One advantage of modeling $W$ on hidden
vector rather than modeling $Q$ on probability
output is that we do not need to project
$Q$ to the space representing  a valid probability transition matrix for each update. 
Another advantage is that 
the linear transition on hidden vectors
can achieve non-linear transition results
on probability space.

\section{Experiments } \label{sec:dataset}

In this section, we demonstrate that our method with
noise converter and conditional optimal selector can reduce the impact of the wrong
labels for relation extraction on distant supervised dataset. We first describe the dataset and the evaluation metrics that are used
in our experiments, and then compare with several baseline methods. 
In addition, we analyze our proposed components by ablation study.

\subsection{Dataset and Evaluation Metrics}

To evaluate our model, we experiment on a widely used dataset which is developed by 
\cite{Riedel_ECML_2010} and has also been tested
by \cite{Hoffmann_ACL_2011,Surdeanu_emnlp_2012,Zeng_emnlp_2015,Lin_ACL_2016}. 
We use the preprocessed version\footnote{\url{https://github.com/thunlp/OpenNRE}} which is made publicly available by Tsinghua NLP Lab.
This dataset was
generated by aligning Freebase relations with the
New York Times corpus (NYT).
The Freebase relations are divided into two parts for training and testing. 
The training set aligns the sentences from the corpus of the years 2005-2006, and the testing one aligns the sentences from 2007. 
The dataset contains 53 possible relationships including a special relation type `NA' which indicates no relation between the mentioned two entities. 
The resulted training and testing data contain 570,088 and 172,448 sentences, respectively. 
We further randomly extract 10 percent of relation pairs and the corresponding sentences from the training data as the validation data for model selection and parameter tuning, and leave the rest as the actual training data. 

Similar to previous works \cite{Mintz_ACL_2009,Lin_ACL_2016},
we evaluate our model in the held-out testing data.
The evaluation compares the extracted relation
instances discovered from the test sentences against Freebase relation data.
It makes the assumption that the inference model
has similar performance in relation instances inside
and outside Freebase. We report the precision/recall curves, Precision@N (P@N), and average precision in our experiments.

\subsection{Parameter Settings}
 We tune the parameters of maximum sentence length,
 learning rate, weight decay, and batch size 
by testing the performance on the validation dataset.
For other parameters, we use the same parameters
as \cite{Lin_ACL_2016}. Table
\ref{tab:para} shows the major parameters used in our experiments.
\vspace*{-0.5cm}
\begin{table}[H]
\caption{Parameter settings.} \label{tab:para}
\begin{center}
	\begin{tabular}{ |c|c|c| } 
		\hline
		Convolution filter window size $l$ & 3 \\
		Number of convolution filters & 230 \\
		Sentence hidden vector  size  & 690 \\
		Word dimension $d^a$ &50  \\ 
		Position dimension $d^b$ & 5  \\ 
		Batch size $B$  & 50  \\ 
		Max sentence length & 100 \\
		Adam learning rate $\lambda$ & 0.001 \\
		Adam weight decay & 0.0001\\ 
		Dropout rate  & 0.5 \\
		\hline
	\end{tabular}
\vspace*{-0.4cm}
\end{center}
\end{table}

For the initialization of $W$, we use the following
strategy. We define a ratio $e$, and assign $w_{11} = 1-e$, and
the rest $K-1$ elements to be $e/ (1-K)$. We do evaluation on the
validation dataset and pick  $e=0.1$ in the candidate set \{0.001, 0.01, 0.1, 0.2, 0.3, 0.5 \}. 
We pretrain our model for 2 epochs by setting $W$ as an identity
matrix and then fine tune our model for another 18 epochs with 
trainable $W$.

\subsection{Comparison with Baseline Methods}

To evaluate our proposed approach, 
we select several baseline methods for comparison
by held-out evaluation:

\textbf{Mintz} \cite{Mintz_ACL_2009}  is a traditional distant
supervised model.

\textbf{MultiR} \cite{Hoffmann_ACL_2011}  is a
probabilistic, graphical model for multi-instance
learning that can handle overlapping relations.

\textbf{MIML} \cite{Surdeanu_emnlp_2012}  is a method that models
both multiple instances and multiple relations.

\textbf{CNN + ATT} and  \textbf{PCNN+ATT} \cite{Lin_ACL_2016} are two methods
that first represent a sentence by CNN and PCNN respectively, and then
use sentence-level selective attention to model a group of sentences with the same entity pair.

For the above baseline methods, we implement 
the state of the art method  \textbf{PCNN+ATT}.  To make fair comparison, we use the same implementation of the component PCNN in our model and  \textbf{PCNN+ATT}, and use the same hyper-parameters. For other methods, we use the results from the source code released
by the authors.

For our method and \textbf{PCNN+ATT}, we run 20 epochs in total, and track
the accuracy on the validation dataset. The model is saved for every 200 batches during training. We use the saved model that has the highest accuracy on the validation dataset for making predictions on the testing dataset.

\begin{figure}[tb]
	\centering
	\includegraphics[width=1.1\linewidth]{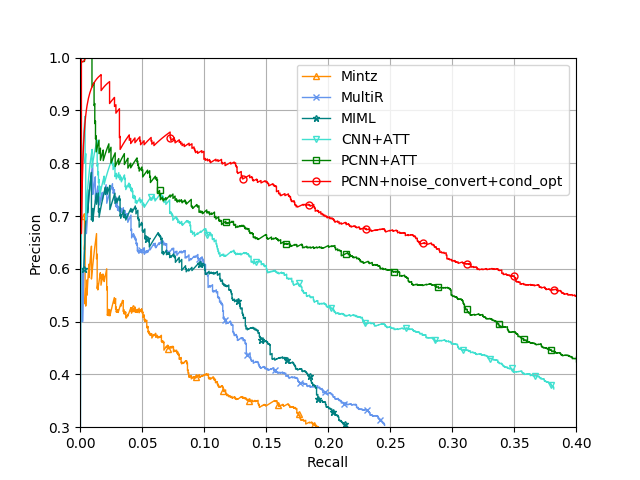}
	\caption{ Performance comparison of proposed
                    model and baseline methods. Our model and our implementation 
                    of PCNN+ATT both pick the model with the highest accuracy on the validation dataset.
	} 
	\label{fig:exp_compare_all}
	\vspace*{-0.4cm}
\end{figure}

Figure \ref{fig:exp_compare_all} shows the precision/recall curves for all methods, 
including ours (labeled as \textbf{PCNN+noise\_convert+cond\_opt}). 
For all of the baseline methods, we can see that PCNN+ATT shows much better performance than others, which demonstrates the effectiveness of the sentence-level selective attention. 
Although PCNN+ATT has shown significant improvement over other baselines, our method still gains great improvement over PCNN+ATT. 
Particularly, Table \ref{tab:p_n_noise_vs_pcnnatt} compares the precision@N (P@N) between our model and PCNN+ATT. 
For PCNN+ATT, we report both the P@N numbers from the authors' original paper and the results based on our implementation.
Our method achieves the highest values for P@100, P@200, P@300, with mean value of 9.1 higher than original report of PCNN+ATT, and 7.1 higher than our implementation of PCNN+ATT. 
%
 %
%
\begin{figure}[tb]
	\centering
	\includegraphics[width=1.1\linewidth]{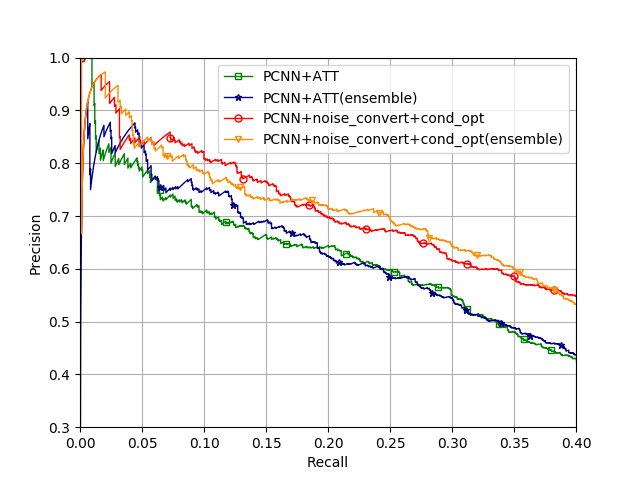}
	\caption{ Performance comparison of proposed
		model and PCNN+ATT model, and their corresponding ensemble versions.
	} 
	\label{fig:exp_compare_assemble}
	\vspace*{-0.4cm}
\end{figure}
%
To avoid randomness in the best single model, we also compare our method and 
PCNN+ATT on ensemble of several saved models in one single training run. 
For each method, the corresponding ensemble model averages the probability scores for each
test instance from the 5 last saved trained single models, and use the average score for prediction. 
Figure \ref{fig:exp_compare_assemble} shows the precision/recall
curves of the corresponding methods with or without ensemble. 
We observe that the performance of the ensemble method of our method is still much better
than the ensemble method of PCNN+ATT. 
We further compare the average precision of our method and PCNN+ATT on both single model and ensemble model, achieving 7.0 and 7.3 higher average precision scores respectively.  
\begin{table}[tb]
	\caption{Comparison of P@N  for relation extraction for our model PCNN+nc+cond\_opt (same notation as  PCNN+noise\_convert+cond\_opt) and PCNN+ATT}  \label{tab:p_n_noise_vs_pcnnatt}
\begin{tabular}{|c|c|c|c|c|c|}
	\hline
	\multicolumn{2}{|c|}{P@N (\%)}  & 100 &200 &300 & Mean \\
	\hline
	\multirow{2}{*}{\shortstack{PCNN \\ +ATT}} & \shortstack{original  report} & 76.2 & 73.1 & 67.4 & 72.2\\
	\cline{2-6}
	& our implementation  & 81.0 & 72.5 & 69 & 74.2  \\
	\cline{2-6}
	\hline
	\multicolumn{2}{|c|}{PCNN+nc+cond\_opt}  & \textbf{85.0} & \textbf{82.0} & \textbf{77.0} & \textbf{81.3} \\
	\hline
\end{tabular}
\end{table}
%
\begin{table}[tb]
	\caption{Comparison of average precision for relation extraction for our model and PCNN+ATT model}  \label{tab:ap_noise_vs_pcnnatt}
	\begin{tabular}{|c|c|c|c|c|c|}
		\hline
		Average Precision (\%)  & Single model & Ensemble model \\
		\hline
		PCNN+ATT  & 36.5 & 37.9 \\		
		\hline
		PCNN+nc+cond\_opt & \textbf{43.5} & \textbf{45.2}  \\
		\hline
	\end{tabular}
\vspace*{-0.4cm}
\end{table}
%
%
\subsection{Ablation Study}

\begin{figure}[tb]
	\centering
	\includegraphics[width=1.1\linewidth]{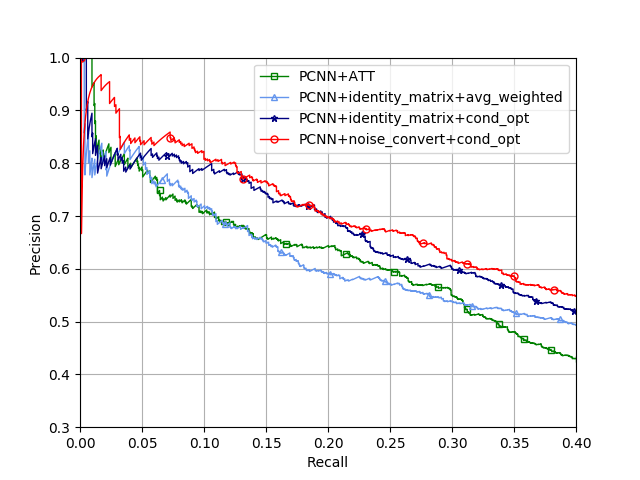}
	\caption{ Performance comparison of PCNN+ATT model
		and our models with different components.
	} 
	\label{fig:exp_impact_of_noise_and_weight}
	\vspace*{-0.4cm}
\end{figure}

To understand the impact of the neural noise converter and conditional optimal selector in our model, we do two more experiments. 
The first one is to  enforce the transition matrix in our model to be identity matrix, and keep other components and loss function remain unchanged, in order to validate the neural noise converter. 
We call this method as \textbf{PCNN+identity\_matrix+cond\_opt } (or PCNN+$\mathbf{I}_W$+cond\_opt ). 
Figure \ref{fig:exp_impact_of_noise_and_weight} illustrates the precision/recall curves of all comparison methods, and Table  \ref{tab:p_n_noise_vs_weight_pcnnatt} shows the P@N values. We can see that without neural noise converter, our method is still better than the baseline method PCNN+ATT, but with neural noise converter, our method gains significantly more improvement. 

In addition, we conduct the second experiment to compare our conditional optimal selector with 
another selector which 
uses the average label distribution of all sentences within each group, called \textbf{PCNN+identity\_matrix+avg\_weighted} (or PCNN+$\mathbf{I}_W$+avg\_weighted ). In practice, we only consider averaging the distribution in the second condition of Equation (\ref{eq:weight}). In other words, this method uses the average probability of all positive predicted sentences. The results also show in Figure \ref{fig:exp_impact_of_noise_and_weight} and Table   \ref{tab:p_n_noise_vs_weight_pcnnatt}.
We can see that this method is slightly better than PCNN+ATT ,
but worse than PCNN+$\mathbf{I}_W$+cond\_opt.
%
%
%
%
\begin{table}[tb]
	\caption{Comparison of P@N  for PCNN+ATT model, and our model with avg\_weighted, cond\_opt and our model with both cond\_opt  and noise\_convert}  \label{tab:p_n_noise_vs_weight_pcnnatt}
	\begin{tabular}{|c|c|c|c|c|c|}
		\hline
		\multicolumn{2}{|c|}{P@N (\%)}  & 100 &200 &300 & Mean \\
		\hline
		\multirow{2}{*}{\shortstack{PCNN \\ +ATT}} & \shortstack{original  report} & 76.2 & 73.1 & 67.4 & 72.2\\
		\cline{2-6}
		& our implementation  & 81.0 & 72.5 & 69 & 74.2  \\
		\cline{2-6}
		\hline
		\multicolumn{2}{|c|}{PCNN + $\mathbf{I}_W$+avg\_weighted}   & 83.0 & 74.5  & 68.3 & 75.2 \\
		\hline
		\multicolumn{2}{|c|}{PCNN + $\mathbf{I}_W$+cond\_opt}   & 81.0 & 78.5  & \textbf{77.0} & 78.8 \\
		\hline
		\multicolumn{2}{|c|}{PCNN+nc+cond\_opt }  & \textbf{85.0} & \textbf{82.0} & \textbf{77.0} & \textbf{81.3} \\
		\hline
	\end{tabular}
\vspace*{-0.4cm}
\end{table}

\section{Conclusions} \label{twitter_sec:conclude}

In this paper, we develop a novel model by
incorporating neural noise converter and
conditional optimal selector to a variant of convolutional
neural network for distantly supervised relation extraction.
We evaluate our
model on the distantly supervised relation extraction task. The experimental results demonstrate that our model significantly and consistently outperforms state-of-the-art methods. 
One possible future work is to relax the constraint of $W^*$ such that our method can be applied to a more general framework, and benefit other NLP tasks. 

\bibliography{mynlp}

\begin{thebibliography}{}

\bibitem[\protect\citeauthoryear{Adel, Roth, and
  Sch{\"{u}}tze}{2016}]{Adel_naacl_2016}
Adel, H.; Roth, B.; and Sch{\"{u}}tze, H.
\newblock 2016.
\newblock Comparing convolutional neural networks to traditional models for
  slot filling.
\newblock In {\em {NAACL} {HLT} 2016, The 2016 Conference of the North American
  Chapter of the Association for Computational Linguistics: Human Language
  Technologies, San Diego California, USA, June 12-17, 2016},  828--838.

\bibitem[\protect\citeauthoryear{Cai, Zhang, and Wang}{2016}]{Cai_ACL_2016}
Cai, R.; Zhang, X.; and Wang, H.
\newblock 2016.
\newblock Bidirectional recurrent convolutional neural network for relation
  classification.
\newblock In {\em Proceedings of the 54th Annual Meeting of the Association for
  Computational Linguistics, {ACL} 2016}.

\bibitem[\protect\citeauthoryear{Fang and Cohn}{2016}]{Fang_CONLL_2016}
Fang, M., and Cohn, T.
\newblock 2016.
\newblock Learning when to trust distant supervision: An application to
  low-resource {POS} tagging using cross-lingual projection.
\newblock In {\em Proceedings of the 20th {SIGNLL} Conference on Computational
  Natural Language Learning, CoNLL 2016},  178--186.

\bibitem[\protect\citeauthoryear{Hoffmann \bgroup et al\mbox.\egroup
  }{2011}]{Hoffmann_ACL_2011}
Hoffmann, R.; Zhang, C.; Ling, X.; Zettlemoyer, L.~S.; and Weld, D.~S.
\newblock 2011.
\newblock Knowledge-based weak supervision for information extraction of
  overlapping relations.
\newblock In {\em The 49th Annual Meeting of the Association for Computational
  Linguistics, {ACL} 2011},  541--550.

\bibitem[\protect\citeauthoryear{Ji \bgroup et al\mbox.\egroup
  }{2017}]{Ji_AAAI_2017}
Ji, G.; Liu, K.; He, S.; and Zhao, J.
\newblock 2017.
\newblock Distant supervision for relation extraction with sentence-level
  attention and entity descriptions.
\newblock In {\em Proceedings of the Thirty-First {AAAI} Conference on
  Artificial Intelligence, 2017},  3060--3066.

\bibitem[\protect\citeauthoryear{Jiang \bgroup et al\mbox.\egroup
  }{2016}]{Jiang_Coling_2016}
Jiang, X.; Wang, Q.; Li, P.; and Wang, B.
\newblock 2016.
\newblock Relation extraction with multi-instance multi-label convolutional
  neural networks.
\newblock In {\em {COLING} 2016, 26th International Conference on Computational
  Linguistics, Proceedings of the Conference},  1471--1480.

\bibitem[\protect\citeauthoryear{Lin \bgroup et al\mbox.\egroup
  }{2016}]{Lin_ACL_2016}
Lin, Y.; Shen, S.; Liu, Z.; Luan, H.; and Sun, M.
\newblock 2016.
\newblock Neural relation extraction with selective attention over instances.
\newblock In {\em Proceedings of the 54th Annual Meeting of the Association for
  Computational Linguistics, {ACL} 2016}.

\bibitem[\protect\citeauthoryear{Luo \bgroup et al\mbox.\egroup
  }{2017}]{Luo_ACL_2017}
Luo, B.; Feng, Y.; Wang, Z.; Zhu, Z.; Huang, S.; Yan, R.; and Zhao, D.
\newblock 2017.
\newblock Learning with noise: Enhance distantly supervised relation extraction
  with dynamic transition matrix.
\newblock In {\em Proceedings of the 55th Annual Meeting of the Association for
  Computational Linguistics, {ACL} 2017},  430--439.

\bibitem[\protect\citeauthoryear{Mintz \bgroup et al\mbox.\egroup
  }{2009}]{Mintz_ACL_2009}
Mintz, M.; Bills, S.; Snow, R.; and Jurafsky, D.
\newblock 2009.
\newblock Distant supervision for relation extraction without labeled data.
\newblock In {\em Proceedings of the Joint Conference of the 47th Annual
  Meeting of the ACL and the 4th International Joint Conference on Natural
  Language Processing of the AFNLP}, ACL '09,  1003--1011.

\bibitem[\protect\citeauthoryear{Misra \bgroup et al\mbox.\egroup
  }{2016}]{Misra_CVPR_2016}
Misra, I.; Zitnick, C.~L.; Mitchell, M.; and Girshick, R.~B.
\newblock 2016.
\newblock Seeing through the human reporting bias: Visual classifiers from
  noisy human-centric labels.
\newblock In {\em CVPR 2016},  2930--2939.

\bibitem[\protect\citeauthoryear{Nguyen and Grishman}{2015}]{Nguyen_NAACL_2015}
Nguyen, T.~H., and Grishman, R.
\newblock 2015.
\newblock Relation extraction: Perspective from convolutional neural networks.
\newblock In {\em Proceedings of the 1st Workshop on Vector Space Modeling for
  Natural Language Processing, VS@NAACL-HLT 2015}.

\bibitem[\protect\citeauthoryear{Reed \bgroup et al\mbox.\egroup
  }{2014}]{Reed_CoRR_2014}
Reed, S.~E.; Lee, H.; Anguelov, D.; Szegedy, C.; Erhan, D.; and Rabinovich, A.
\newblock 2014.
\newblock Training deep neural networks on noisy labels with bootstrapping.
\newblock {\em CoRR} abs/1412.6596.

\bibitem[\protect\citeauthoryear{Riedel, Yao, and
  McCallum}{2010}]{Riedel_ECML_2010}
Riedel, S.; Yao, L.; and McCallum, A.
\newblock 2010.
\newblock Modeling relations and their mentions without labeled text.
\newblock In {\em Machine Learning and Knowledge Discovery in Databases,
  European Conference, {ECML} {PKDD} 2010},  148--163.

\bibitem[\protect\citeauthoryear{Santos, Xiang, and
  Zhou}{2015}]{Santos_ACL_2015}
Santos, C. N.~D.; Xiang, B.; and Zhou, B.
\newblock 2015.
\newblock Classifying relations by ranking with convolutional neural networks.
\newblock In {\em Proceedings of the 53rd Annual Meeting of the Association for
  Computational Linguistics and the 7th International Joint Conference on
  Natural Language Processing of the Asian Federation of Natural Language
  Processing, (ACL) 2015}.

\bibitem[\protect\citeauthoryear{Sukhbaatar \bgroup et al\mbox.\egroup
  }{2015}]{Sukhbaatar_ICLR_2015}
Sukhbaatar, S.; Bruna, J.; Paluri, M.; Bourdev, L.; and Fergus, R.
\newblock 2015.
\newblock Training convolutional networks with noisy labels.
\newblock {\em {ICLR}}.

\bibitem[\protect\citeauthoryear{Surdeanu \bgroup et al\mbox.\egroup
  }{2012}]{Surdeanu_emnlp_2012}
Surdeanu, M.; Tibshirani, J.; Nallapati, R.; and Manning, C.~D.
\newblock 2012.
\newblock Multi-instance multi-label learning for relation extraction.
\newblock In {\em Proceedings of the 2012 Joint Conference on Empirical Methods
  in Natural Language Processing and Computational Natural Language Learning,
  EMNLP-CoNLL 2012},  455--465.

\bibitem[\protect\citeauthoryear{Takamatsu, Sato, and
  Nakagawa}{2012}]{Shingo_ACL_2012}
Takamatsu, S.; Sato, I.; and Nakagawa, H.
\newblock 2012.
\newblock Reducing wrong labels in distant supervision for relation extraction.
\newblock In {\em The 50th Annual Meeting of the Association for Computational
  Linguistics, Proceedings of the Conference},  721--729.

\bibitem[\protect\citeauthoryear{Wang \bgroup et al\mbox.\egroup
  }{2016}]{Wang_ACL_2016}
Wang, L.; Cao, Z.; de~Melo, G.; and Liu, Z.
\newblock 2016.
\newblock Relation classification via multi-level attention cnns.
\newblock In {\em Proceedings of the 54th Annual Meeting of the Association for
  Computational Linguistics, {ACL} 2016}.

\bibitem[\protect\citeauthoryear{Xu \bgroup et al\mbox.\egroup
  }{2013}]{Xu_ACL_2013}
Xu, W.; Hoffmann, R.; Zhao, L.; and Grishman, R.
\newblock 2013.
\newblock Filling knowledge base gaps for distant supervision of relation
  extraction.
\newblock In {\em Proceedings of the 51st Annual Meeting of the Association for
  Computational Linguistics, {ACL} 2013},  665--670.

\bibitem[\protect\citeauthoryear{Zeng \bgroup et al\mbox.\egroup
  }{2014}]{Zeng_coling_2014}
Zeng, D.; Liu, K.; Lai, S.; Zhou, G.; and Zhao, J.
\newblock 2014.
\newblock Relation classification via convolutional deep neural network.
\newblock In {\em {COLING} 2014, 25th International Conference on Computational
  Linguistics, Proceedings of the Conference: Technical Papers, 2014},
  2335--2344.

\bibitem[\protect\citeauthoryear{Zeng \bgroup et al\mbox.\egroup
  }{2015}]{Zeng_emnlp_2015}
Zeng, D.; Liu, K.; Chen, Y.; and Zhao, J.
\newblock 2015.
\newblock Distant supervision for relation extraction via piecewise
  convolutional neural networks.
\newblock In {\em Proceedings of the 2015 Conference on Empirical Methods in
  Natural Language Processing, {EMNLP} 2015},  1753--1762.

\bibitem[\protect\citeauthoryear{Zhou \bgroup et al\mbox.\egroup
  }{2016}]{Zhou_ACL_2016}
Zhou, P.; Shi, W.; Tian, J.; Qi, Z.; Li, B.; Hao, H.; and Xu, B.
\newblock 2016.
\newblock Attention-based bidirectional long short-term memory networks for
  relation classification.
\newblock In {\em Proceedings of the 54th Annual Meeting of the Association for
  Computational Linguistics, {ACL} 2016, August 7-12, 2016, Berlin, Germany,
  Volume 2: Short Papers}.

\end{thebibliography}

\bibliographystyle{aaai}



\end{document}